\def\year{2021}\relax
\documentclass[letterpaper]{article} 
\usepackage{aaai21}  
\usepackage{times}  
\usepackage{helvet} 
\usepackage{courier}  
\usepackage[hyphens]{url}  
\usepackage{graphicx} 
\usepackage{natbib}  
\usepackage{caption} 
\usepackage{amsmath}
\usepackage{multirow}
\usepackage{multicol}
\usepackage{booktabs}
\usepackage{color}
\usepackage{hyperref}
\usepackage{cuted}

\title{Multi-Decoder Attention Model with Embedding Glimpse for Solving Vehicle Routing Problems}
\author{
    Liang Xin,\textsuperscript{\rm 1}\footnotemark[1] Wen Song,\textsuperscript{\rm 2}\thanks{Liang Xin and Wen Song contributed equally.} Zhiguang Cao,\textsuperscript{\rm 3}\thanks{Zhiguang Cao is the corresponding author.} Jie Zhang\textsuperscript{\rm 1}
}
\affiliations{
    \\
    \textsuperscript{\rm 1}Nanyang Technological University,
    \textsuperscript{\rm 2}Shandong University,
    \textsuperscript{\rm 3}National University of Singapore
    \\
    XINL0003@e.ntu.edu.sg, 
    wensong@email.sdu.edu.cn,
    zhiguangcao@outlook.com,
    zhangj@ntu.edu.sg

}

\begin{document}
\maketitle

\begin{abstract}
We present a novel deep reinforcement learning method to learn construction heuristics for vehicle routing problems. In specific, we propose a Multi-Decoder Attention Model (MDAM) to train multiple diverse policies, which effectively increases the chance of finding good solutions compared with existing methods that train only one policy. A customized beam search strategy is designed to fully exploit the diversity of MDAM. In addition, we propose an Embedding Glimpse layer in MDAM based on the recursive nature of construction, which can improve the quality of each policy by providing more informative embeddings. Extensive experiments on six different routing problems show that our method significantly outperforms the state-of-the-art deep learning based models.
\end{abstract}

\section{Introduction}

Routing problems, such as the Travelling Salesman Problem (TSP) and Capacitated Vehicle Routing Problem (CVRP), are a family of combinatorial optimization problems (COP) that have extensive real-world applications in many domains \cite{toth2014vehicle}. Due to the combinatorial nature, routing problems are NP-hard in general \cite{applegate2006traveling}. Exact approaches, such as branch-and-bound algorithms \cite{fischetti1994branch}, have nice theoretical guarantee of optimality, but the (worst-case) computation complexity is exponential. 
In contrast, approximate algorithms guided by  heuristics can find near-optimal solutions with polynomial computation complexity, therefore are often preferred, especially for large-scale problems.

Traditional approaches design hand-crafted rules as the heuristics. Instead, as modern approaches, deep learning models learn the heuristics from data samples
\cite{bello2016neural,dai2016discriminative,nazari2018reinforcement,kool2018attention,chen2019learning}.
Most of these deep learning methods follow the encoder-decoder structure, and learn construction heuristics by repeatedly adding nodes (or locations) into an empty or partial solution until completion. Particularly, the encoder maps the information of nodes into feature embeddings, and the decoder predicts the probabilities of selecting each valid node at every construction step. To improve solution quality, different methods (e.g. sampling \cite{kool2018attention} or beam searching \cite{nazari2018reinforcement}) are used to generate a set of solutions from the trained construction policy to get the best one.

Though showing promising results, existing works suffer from two major limitations. First, the generated solutions are not \emph{diverse} enough. Intuitively, a more diverse set of solutions could potentially lead to better ones. This is because for VRP and many other COPs, multiple optimal solutions exist and trying to find different ones will increase the chance of finding at least one. In addition, with the same number of solutions, generating less diverse and partially identical solutions will leave less space for the potentially better ones. Existing methods train only one constructive policy discriminatively, and the solutions are created using sampling or beam search from this same policy. The only source of diversity comes from the relatively deterministic probability distribution, which is far from enough. 
The second limitation, as pointed out in \cite{9226142}, is regarding the training of construction policy itself. The construction process can be viewed as a sequence of node selection sub-tasks, where the already visited node is \emph{irrelevant} to the future decisions. However, most existing models for learning construction heuristic \cite{kool2018attention,bello2016neural} use the same node embeddings to decode at each step, without eliminating irrelevant nodes. Therefore, the node embeddings are based on the original graph for the whole task, not the graphs for each sub-task, and may deteriorate the quality of the trained policy.

In this paper, we address the above limitations simultaneously. First, to improve diversity, we propose the Multi-Decoder Attention Model (MDAM) to train multiple construction policies. It employs a Transformer \cite{vaswani2017attention} to encode the node information, and multiple identical attention decoders with unshared parameters to sample different trajectories. During training, each of the decoders learns distinct solution patterns, and is regularized by a Kullback-Leibler divergence loss to force the decoders to output dissimilar probability distribution of selecting nodes. Based on MDAM, we propose a novel beam search scheme where separate beams are maintained for each decoder. This enables full utilization of the distinct patterns learned by each decoder, and effectively keeps the diversity of solutions. 
Secondly, to increase the quality of trained construction policies, we propose an Embedding Glimpse layer in MDAM by exploiting the recursive nature of routing problems. As the visited nodes become unrelated to the future decisions, we explicitly remove them in the top attention layer of our encoder. Therefore the decoders will get more informative embeddings for selecting the next node,
hence increase the quality of each single solution.

We would like to note that, rather than outperforming those highly optimized solvers in general, we are here to push the edge of deep learning model towards learning stronger heuristics for routing problems and potentially other combinatorial problems with weak hand-designed heuristics. Similar to \cite{kool2018attention}, while focusing on TSP and CVRP, our method is flexible and generally applicable to a wide range of routing problems with different constraints and even uncertainty. Extensive experiment results on six routing problems well confirm the effectiveness of increasing diversity and removing irrelevant nodes. More importantly, our model significantly outperforms state-of-the-art deep reinforcement learning based methods, and also demonstrates comparable or superior performance to the traditional non-learning based heuristics and sophisticated solvers in short inference time.

\section{Related Works}
Among existing models for learning construction heuristics, Pointer Network (PtrNet) in \cite{vinyals2015pointer} uses Long Short-Term Memory (LSTM) Networks ~\cite{hochreiter1997long} as encoder and decoder to solve TSP with supervised learning, hence limited to small scale due to the expensive query for labels (optimal solutions).
In contrast, \citeauthor{bello2016neural} \shortcite{bello2016neural} use REINFORCE algorithm \cite{williams1992simple} to train PtrNet. Without the need for true labels, the model can be trained on TSP with larger sizes.
Instead of encoding nodes sequentially with LSTM in PtrNet, \citeauthor{nazari2018reinforcement} \shortcite{nazari2018reinforcement} use permutation invariant layers to encode nodes and train this model for CVRP.
They also improve the solution quality using beam search to keep track of the most promising solutions and choose the best one. \citeauthor{kool2018attention} \shortcite{kool2018attention} adopt the Transformer model \cite{vaswani2017attention} to encode nodes and use a pointer-like attention mechanism to decode. By sampling 1,280 solutions from the trained policy, this elegant model achieves state-of-the-art results on several routing problems.

Some other works do not adopt the above encoder-decoder structure. \citeauthor{khalil2017learning} \shortcite{khalil2017learning} use Deep Q-Learning algorithm to train a deep architecture over graph, i.e. Structure2Vec \cite{dai2016discriminative}, for several combinatorial optimization problems including TSP. However, the full connectivity makes the graph structure of TSP unimportant and results in unsatisfactory performance compared to that of other problems with crucial graph information, such as Minimum Vertex Cover and Maximum Cut.
Instead of learning construction heuristics, \citeauthor{chen2019learning} \shortcite{chen2019learning} propose NeuRewriter to learn improvement heuristics, and train a region-picking policy and a rule-picking policy that recursively refine an initial solution for certain steps. For CVRP, NeuRewriter outperforms the sampling results in \cite{kool2018attention}. However, unlike our easily parallelized searching method, NeuRewriter is naturally not parallelizable when solving an instance since it uses sequential rewriting operations. Different from learning to pick nodes in construction or improvement heuristics, \citeauthor{iclr2020} \shortcite{iclr2020} design a network to learn which type of local move to pick and exhaustively search for the best possible greedy move of this type at each local improvement step. Despite the good solution quality, this model is not practical due to its prohibitively long computation time caused by the tens of thousands of exhaustive searching over the local move space at each step. 

\section{Model}

\begin{figure*}[t]
\centering
\includegraphics[width=0.97\textwidth]{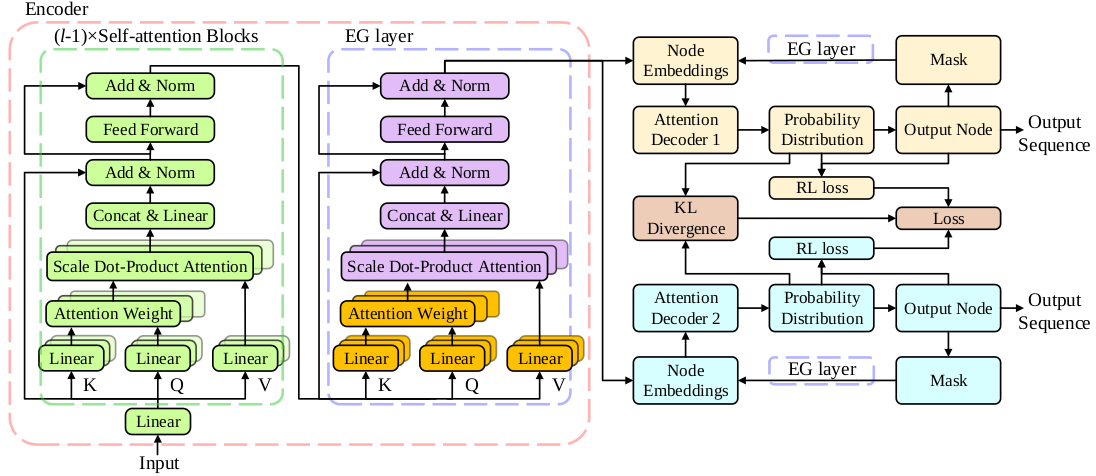}
\caption{The Multi-Decoder Attention Model (we use two decoders for illustration)}
\label{MDAM}
\end{figure*}

\subsection{Multi-Decoder Attention Model and Search}
To output diverse solutions, we propose the Multi-Decoder Attention Model (MDAM) and design a customized beam search scheme utilizing the structure of multiple decoders to effectively keep the diversity of the solutions in the beam.
\subsubsection{Multi-Decoder Attention Model}
The MDAM consists of an encoder and multiple decoders with identical structures but unshared parameters. The encoder takes the input instance $x$ as the two-dimensional coordinates of nodes in TSP
and embed them into feature vectors.
For other routing problems, additional dimensions such as \emph{demands} and \emph{prizes} are included as input.
During each step of the solution construction, each decoder takes node embeddings to produce probabilities of visiting each valid node. The MDAM architecture is shown in Figure \ref{MDAM}, while EG layer will be detailed later.

The encoder of MDAM follows the Transformer Model \cite{vaswani2017attention}. It consists of multiple self-attention blocks, the core of which is the multi-head attention layer and is defined formally as follows:

\begin{equation} \label{mh1}
    Q^h_i,K^h_i,V^h_i=W^h_{Q}X_i,W^h_{K}X_i,W^h_{V}X_i,
\end{equation}
\begin{equation} \label{mh2}
\begin{split}
    A^h  &= \text{Attention}(Q^h,K^h,V^h) \\
    &= \text{softmax}(Q^h{K^h}^T/\sqrt{d_k})V^h, h=1,2,...,H,
\end{split}
\end{equation}
\begin{equation}
    \text{Multihead}(Q,K,V)=\text{Concat}(A^1,A^2,...,A^H)W_O,
\end{equation}
where Eqs.~({\ref{mh1}) and (\ref{mh2}}) are performed for each of the $H$ attention heads; $X_i$ is the $d$-dimensional embedding for the ith node; $Q, K, V$ are \emph{Query}, \emph{Key}, \emph{Value} vectors, respectively, $W^h_Q, W^h_K, W^h_V\in R^{d\times d_k}$ with $d_k=d/H$; the attention outputs $A^h$ are concatenated and projected with $W_O\in R^{d\times d}$ to get this multi-head attention layer output.

Then skip-connection layers~\cite{he2016deep}, Batch Normalization (BN) layers~\cite{ioffe2015batch} and two linear projection layers with ReLU activation in between (referred as FF) are used to get the output $f$ of this self-attention block as follows:
\begin{equation} \label{eq101}
    \hat{f_i}=\text{BN}(X_i+\text{Multihead}_i(Q,K,V)),
\end{equation}
\begin{equation} \label{eq102}
    f_i=\text{BN}(\hat{f_i}+\text{FF}(\hat{f_i})).
\end{equation}

Let $M$ be the number of decoders with identical structure. Each decoder, indexed by $m$, is an attention mechanism that models the probability of selecting next node to visit at each step $t$, $P^m(y_t|x, y_1, ..., y_{t-1})$, following \cite{kool2018attention}. 
The decoder indexed by $m$ is defined formally as follows:
\begin{equation}
    f_{c}=\text{Concat}(\bar{f},f_{C_0},f_{C_{t-1}}),    
\end{equation}
\begin{equation} \label{eq16}
    g^m_{c}=\text{Multihead}(W^m_{gQ}f_{c}, W^m_{gK}f, W^m_{gV}f),
\end{equation}
\begin{equation}
    q^m,k^m_i=W^m_Qg^m_{c},W^m_Kf_i,    
\end{equation}
\begin{equation} \label{tanh}
    u^m_i=D\tanh((q^m)^Tk^m_i/\sqrt{d}),
\end{equation}
\begin{equation} \label{softmax1}
    P^m(y_t|x, y_1, ..., y_{t-1})=\text{softmax}(u^m),
\end{equation}
where $f_{c}$ is the context embedding; $\bar{f}$ is the mean of the nodes' embeddings $f$; $f_{C_0}$ and $f_{C_{t-1}}$ are the embeddings of the starting node and the current node, respectively, which are replaced by trainable parameters for the first step; Eq.~(\ref{eq16}) is a mulit-head attention over the valid nodes at step $t$ to get a new context, similar to the glimpse in \cite{bello2016neural}; $g^m_{c},q^m,k^m_i$ are of dimension $d$; Eq.~(\ref{tanh}) uses $D=10$ to clip the result for better exploration following \cite{bello2016neural}; the softmax function in Eq.~(\ref{softmax1}) is over the set of valid nodes.

While greedily decoding, each decoder independently outputs a trajectory by selecting the node with maximum probability at each step. We impose a regularization to encourage the decoders to learn distinct construction patterns and output diverse solutions. A Kullback-Leibler (KL) divergence between each pair of the output probability distributions from the multiple decoders of MDAM is maximized as the regularization during training:
\begin{equation}
    D_{KL}=\sum_s\sum_{i=1}^M\sum_{j=1}^M\sum_{y}P^i(y|x,s)\log\frac{P^i(y|x,s)}{P^j(y|x,s)},
\end{equation}
where $P^i(y|x, s)$ is the probability that decoder $i$ selects node $y$ in state $s$ for instance $x$.

\subsubsection{Customized Beam Search}

While the solution space for combinatorial optimization problems is exponentially large, evaluating a small set of solutions is computationally feasible with the known deterministic metric (e.g. tour length for routing problems). Therefore, it is desirable to output a set of solutions and retrieve the best one. However, with learned construction heuristics, existing methods like the typical sampling and beam search fail to maintain a set of diverse solutions. This issue comes from the fact that randomly sampling (with replacements) from the same distribution (policy) frequently outputs partially or completely repeated solutions, and beam search on a tree finds a set of unique solutions but with low variability and being deterministic \cite{kool2019stochastic}.

To achieve better performance, we propose a novel beam search scheme customized for MDAM where each decoder performs search independently. Given a required beam size $\mathcal{B}$, we maintain a separate beam for each decoder to keep the diversity of the whole beam, and the size of a beam is $B=\lceil \mathcal{B}/M \rceil$. The solutions in the same beam will utilize a consistent construction pattern since they are constructed by the same decoder.

\begin{figure}[t]
\centering
\includegraphics[width=0.4\textwidth]{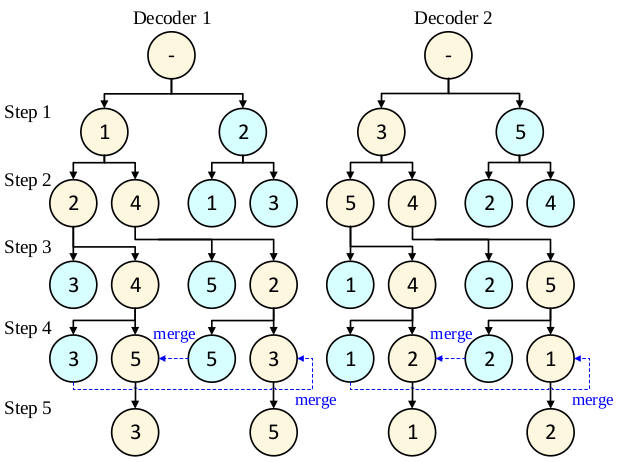}
\caption{The Multi-Decoder Attention Model (we use two decoders for illustration)}
\label{search_fig}
\end{figure}

During searching of the same decoder, we can usually conclude the \emph{inferiority} of a partial solution to another, without the need of full construction. Taking TSP for example, some partial solutions of a decoder may collapse into having the same starting node, set of visited nodes, and current node, i.e. the same remaining sub-task.
Figure \ref{search_fig} depicts an example, where the second (1$\to$2$\to$4$\to$5) and third (1$\to$4$\to$2$\to$5) partial solutions in step 4 of decoder 1 collapse. One of these two will have a partial tour length longer than or equal to the other. Hence with all possible future decisions of the same decoder, it can never achieve shorter complete tour length, i.e. inferior to the other one. For CVRP, similar analysis also applies. But to ensure inferiority, in addition to the collapse condition, the partial solution with the tour length longer than or equal to the other should have less or same vehicle capacity left.

Here we design a merging technique based on the inferiority conditions above to avoid unnecessary solution constructions and save space in the beam for potentially better ones. Specifically, in each beam we evaluate partial solutions on the fly, and merge them whenever one is inferior to another one.
After merging, the inferior one will be deleted, and the probability of the superior one will take the larger value of the merged two. Due to pairwise comparisons, the computation overhead for merging grows quadratically with the increase of beam size. Nevertheless, this is not a critical issue for our method since we maintain separate beams for each decoder, the sizes of which ($B$) are often small.

\subsection{Embedding Glimpse Layer} \label{EG}

For construction heuristics, the model needs to output a series of node selections in an auto-regressive way. The already visited nodes are no longer relevant to the future decisions, and decoding based on the same embeddings for all the construction steps may lead to poor performance. To further improve the solution quality, an ideal way is to  re-embed only the unvisited nodes after visiting one node at each tour construction step.
However, this requires complete re-embedding of nodes in each step, and is extremely expensive. To address it, we propose the Embedding Glimpse (EG) layer to approximate the re-embedding process. This is based on the intuition that higher layer of a neural network extracts more task-related features. 
Taking the Convolutional Neural Network (CNN) for Computer Vision tasks as an example, the lower layers usually extract general features like pixel and edge patterns, while the features from the upper layers are more related to solving the specific tasks \cite{schroff2015facenet}.
We hypothesize that the attention model shares a similar hierarchical feature extraction paradigm, and design the EG layer as follows.

In the encoder with $l$ self-attention blocks,
we perform the re-embedding process approximately by fixing the lowest $l-1$ attention layers
and masking out the attention weights of the irrelevant nodes in the top attention layer to get the new node embeddings.
Termed as EG layer, this top attention layer is part of the encoder and keeps the same for each of the multiple decoders.
Part of the EG layer can be pre-computed for only one time (the orange boxes in Figure \ref{MDAM}):
\begin{equation}
    Q^h_i,K^h_i,V^h_i=W^h_{Q}X_i,W^h_{K}X_i,W^h_{V}X_i,
\end{equation}
\begin{equation}
    w^h(Q^h,K^h)=Q^h{K^h}^T/\sqrt{d_k}.
\end{equation}
At step $t$, we mask out the attention weights to the visited nodes $C_{t-1}$ by setting $w^h_{\cdot,C_{t-1}}=-\infty $ and do the following (the purple boxes in Figure\ref{MDAM}):
\begin{equation} \label{eq15}
    A^h=\text{Attention}(Q^h,K^h,V^h)=\text{softmax}(w^h)V^h,
\end{equation}
\begin{equation}\label{eq99}
    \text{Multihead}(Q,K,V)=\text{Concat}(A^1,A^2,...,A^H)W_O.
\end{equation}
Then Eqs.~(\ref{eq101}) and (\ref{eq102}) engender the node embeddings, 
based on which each decoder selects a node to visit and gets a new mask. After that, while making a new decision about which node to visit, new node embeddings can be achieved
by performing Eqs.~(\ref{eq15}), (\ref{eq99}), (\ref{eq101}) and (\ref{eq102}) with the new mask. 

The EG layer can be considered as correcting the node embeddings with the information that the visited nodes should be no longer relevant. However, running EG layer every step incurs additional computational overhead. To alleviate this, we perform EG layer every $p$ steps. This is reasonable because the node embeddings change gradually and only one node is removed in a step. We found this technique consistently boosts the performance with little inference time. The EG layer can be viewed as a generalization of the step-wise idea in \cite{9226142} with better computation efficiency. By choosing the hyper-parameter $p$, we can keep the number of times to re-embed with EG layer approximately constant for problems with different sizes.

\subsection{Training}
The MDAM structure and the training process is shown in Figure \ref{MDAM}. The $l$-layer encoder consists of $l$-1 attention blocks and one EG layer which we have introduced before.
For input instance $x$, each of the multiple decoders individually samples a trajectory $\pi_m$ to get separate REINFORCE loss with the same greedy roll-out baseline from MDAM:
\begin{equation}
    \nabla \mathcal{L}_{RL}(\theta | x)=\sum_{m}\displaystyle \mathop{E}_{P^m_\theta(\pi_m|x)}(L(\pi_m)-b(x))\nabla \log P^m_\theta(\pi_m |x),
\end{equation}
where $L(\pi_m)$ is the tour length. The baseline we adopt is similar to the one with the best performance in \cite{kool2018attention}. 
The model with the best set of parameters among previous epochs is used as the baseline model to greedily decode the result as the baseline $b(x)$.
\begin{equation}
    b(x)=\min_{m} L(\pi_m'=\{y_m'(1),...,y_m'(T)\}),
\end{equation}
\begin{equation}
    y_m'(t)=\mathop{\arg\max}_{y_t}(P_{\theta'}^m(y_t|x, y_m'(1),...,y_m'(t-1))),
\end{equation}
where $\theta$ is the current parameters of MDAM, $\theta'$ is the fixed parameters of baseline model from previous epoch, and $m$ is the decoder index.
We optimize the model by gradient descent:
\begin{equation}
    \nabla \mathcal{L}(\theta)=\nabla \mathcal{L}_{RL}(\theta | x)-k_{KL}\nabla D_{KL},
\end{equation}
where $k_{KL}$ is the coefficient of KL loss.
Ideally, for each state encountered by each decoder, a KL loss should be computed during training to encourage diversity. To avoid expensive computation, we only impose KL loss on the first step, motivated by the following reasons. First, the initial state of all decoders is the same, i.e. an empty solution, where KL loss is meaningful and easy to compute since all decoders start from this state. Second, for the same instance, different optimal solutions usually have different options at the first step, which have strong impact on the construction patterns.

\begin{table*}[p] \small
\centering
\resizebox{\textwidth}{!}{
\begin{tabular}{ll|lrr|lrr|lrr}
\toprule
\multicolumn{2}{c|}{\multirow{2}{*}{Method}} & \multicolumn{3}{c|}{n=20} & \multicolumn{3}{c|}{n=50} & \multicolumn{3}{c}{n=100} \\
\multicolumn{2}{c|}{}  & Obj & Gap & Time & Obj & Gap & Time & Obj & Gap & Time  \\
\midrule
\multirow{5}{*}{\rotatebox{90}{TSP}} &Concorde              & 3.84*  & 0.00\% & 1m & 5.70* & 0.00\% & 2m & 7.76* & 0.00\% & 3m      \\
&AM greedy \cite{kool2018attention}        & 3.85  & 0.34\% & 0s & 5.80 & 1.76\% & 2s & 8.12 & 4.53\% & 6s      \\
&AM sampling \cite{kool2018attention}      & 3.84  & 0.08\% & 5m & 5.73 & 0.52\% & 24m & 7.94 & 2.26\% & 1h       \\
&MDAM greedy      & 3.84  & 0.05\% & 5s & 5.73 & 0.62\% & 15s & 7.93 & 2.19\% & 36s      \\
&MDAM bs30   & \textbf{3.84}  & \textbf{0.00\%} & 2m & \textbf{5.70} & \textbf{0.04\%} & 7m & \textbf{7.80} & \textbf{0.48\%} & 20m      \\
&MDAM bs50   & \textbf{3.84}  & \textbf{0.00\%} & 3m & \textbf{5.70} & \textbf{0.03\%} & 14m & \textbf{7.79} & \textbf{0.38\%} & 44m      \\
\midrule
\multirow{8}{*}{\rotatebox{90}{CVRP}} &LKH              & 6.14*  & 0.00\% & 2h & 10.38* & 0.00\% & 7h & 15.65* & 0.00\% & 13h      \\
& RL (beam 10) \cite{nazari2018reinforcement}    & 6.40 & 4.39\% & 27m & 11.15 & 7.46\% & 39m & 16.96 & 8.39\% & 74m \\
&AM greedy \cite{kool2018attention}       & 6.40  & 4.43\% & 1s & 10.98 & 5.86\% & 3s & 16.80 & 7.34\% & 8s      \\
&AM sampling \cite{kool2018attention}      & 6.25  & 1.91\% & 6m & 10.62 & 2.40\% & 28m & 16.23 & 3.72\% & 2h       \\
&NeuRewriter \cite{chen2019learning}     & 6.16  & 0.48\% & 22m & 10.51 & 1.25\% & 35m & 16.10 & 2.88\% & 66m      \\
&MDAM greedy      & 6.24  & 1.79\% & 7s & 10.74 & 3.47\% & 16s & 16.40 & 4.86\% & 45s      \\
&MDAM bs30   & \textbf{6.14}  & \textbf{0.26\%} & 3m & \textbf{10.50} & \textbf{1.18\%} & 9m & \textbf{16.03} & \textbf{2.49\%} & 31m      \\
&MDAM bs50   & \textbf{6.14}  & \textbf{0.18\%} & 5m & \textbf{10.48} & \textbf{0.98\%} & 15m & \textbf{15.99} & \textbf{2.23\%} & 53m      \\
\midrule
\multirow{7}{*}{\rotatebox{90}{SDVRP}} & RL (greedy) \cite{nazari2018reinforcement}    & 6.51 & 5.77\% & - & 11.32 & 8.07\% & - & 17.12 & 7.17\% & - \\
& RL (beam 10) \cite{nazari2018reinforcement}    & 6.34 & 3.01\% & - & 11.08 & 5.78\% & - & 16.86 & 5.54\% & - \\
&AM greedy \cite{kool2018attention}       & 6.39  & 3.82\% & 1s & 10.92 & 4.25\% & 4s & 16.83 & 5.36\% & 11s      \\
&AM sampling \cite{kool2018attention}      & 6.25  & 1.55\% & 9m & 10.59 & 1.10\% & 42m & 16.27 & 1.85\% & 3h       \\
&MDAM greedy      & 6.25  & 1.49\% & 13s & 10.72 & 2.31\% & 28s & 16.39 & 2.62\% & 1m      \\
&MDAM bs30   & \textbf{6.16}  & \textbf{0.08\%} & 4m & \textbf{10.49} & \textbf{0.18\%} & 11m & \textbf{16.01} & \textbf{0.24\%} & 28m      \\
&MDAM bs50   & \textbf{6.15}*  & \textbf{0.00\%} & 6m & \textbf{10.47}* & \textbf{0.00\%} & 19m & \textbf{15.97}* & \textbf{0.00\%} & 1h      \\
\midrule
\multirow{8}{*}{\rotatebox{90}{OP}} & Gurobi \cite{gurobi} & 5.39* & 0.00\% & 16m & \multicolumn{3}{c|}{-} & \multicolumn{3}{c}{-} \\
& Gurobi \cite{gurobi} (30s) & 5.38 & 0.05\% & 14m & 13.57 & 16.29\% & 2h & 3.23 & 90.28\% & 3h \\
& Compass\cite{kobeaga2018efficient} & 5.37 & 0.36\% & 2m & 16.17 & 0.25\% & 5m & 33.19* & 0.00\% & 15m \\
&AM greedy \cite{kool2018attention} & 5.19  & 3.64\% & 0s & 15.64 & 3.52\% & 1s & 31.62 & 4.75\% & 5s      \\
&AM sampling \cite{kool2018attention} & 5.30  & 1.56\% & 4m & 16.07 & 0.87\% & 16m & 32.68 & 1.55\% & 53m       \\
&MDAM greedy & 5.32  & 1.32\% & 7s & 15.92 & 1.80\% & 14s & 32.32 & 2.61\% & 32s      \\
&MDAM bs30   & \textbf{5.38}  & \textbf{0.15\%} & 1m & \textbf{16.19} & \textbf{0.10\%} & 6m & \textbf{32.91} & \textbf{0.84\%} & 14m      \\
&MDAM bs50   & \textbf{5.38}  & \textbf{0.13\%} & 3m & \textbf{16.21}* & \textbf{0.00\%} & 15m & \textbf{32.96} & \textbf{0.69\%} & 32m      \\
\midrule
\multirow{9}{*}{\rotatebox{90}{PCTSP}} & Gurobi \cite{gurobi} & 3.13* & 0.00\% & 2m & \multicolumn{3}{c|}{-} & \multicolumn{3}{c}{-} \\
& Gurobi \cite{gurobi} (30s) & 3.13* & 0.00\% & 2m & 4.48 & 0.03\% & 54m & \multicolumn{3}{c}{-} \\
&OR Tools \cite{google2016} (60s) & 3.13  & 0.01\% & 5h & 4.48* & 0.00\% & 5h & 6.07 & 1.56\% & 5h      \\
& ILS C++ \cite{lourencco2003iterated} & 3.16  & 0.77\% & 16m & 4.50 & 0.36\% & 2h & 5.98* & 0.00\% & 12h      \\
&AM greedy \cite{kool2018attention}       & 3.18  & 1.62\% & 0s & 4.60 & 2.66\% & 2s & 6.25 & 4.46\% & 5s      \\
&AM sampling \cite{kool2018attention}      & 3.15  & 0.45\% & 5m & 4.52 & 0.74\% & 19m & 6.08 & 1.67\% & 1h       \\
&MDAM greedy & 3.16  & 0.82\% & 7s & 4.56 & 1.73\% & 18s & 6.17 & 3.13\% & 34s      \\
&MDAM bs30   & \textbf{3.14}  & \textbf{0.21\%} & 2m & \textbf{4.50} & \textbf{0.55\%} & 9m & \textbf{6.07} & \textbf{1.46\%} & 16m      \\
&MDAM bs50   & \textbf{3.14}  & \textbf{0.19\%} & 4m & \textbf{4.50} & \textbf{0.47\%} & 23m & \textbf{6.06} & \textbf{1.31\%} & 35m      \\
\midrule
\multirow{7}{*}{\rotatebox{90}{SPCTSP}} & REOPT all & 3.34 & 6.05\% & 17m & 4.68 & 3.69\% & 2h & 6.22 & 2.52\% & 12h \\
& REOPT half & 3.31 & 5.10\% & 25m & 4.64 & 2.80\% & 3h & 6.16 & 1.53\% & 16h \\

&AM greedy \cite{kool2018attention}       & 3.26  & 3.51\% & 0s & 4.65 & 3.03\% & 2s & 6.32 & 4.17\% & 5s      \\
&AM sampling \cite{kool2018attention}      & 3.19  & 1.30\% & 2m & 4.54 & 0.60\% & 17m & 6.11 & 0.69\% & 40m       \\
&MDAM greedy & 3.19  & 1.34\% & 9s & 4.59 & 1.61\% & 18s & 6.18 & 1.84\% & 1m      \\
&MDAM bs30   & \textbf{3.15}  & \textbf{0.05\%} & 2m & \textbf{4.52} & \textbf{0.12\%} & 14m & \textbf{6.08} & \textbf{0.16\%} & 26m      \\
&MDAM bs50   & \textbf{3.15}*  & \textbf{0.00\%} & 4m & \textbf{4.51}* & \textbf{0.00\%} & 15m & \textbf{6.07}* & \textbf{0.00\%} & 38m      \\
\bottomrule
\end{tabular}
}
{\raggedright Note: We evaluate our model using a single RTX-2080Ti GPU. AM sampling samples 1,280 solutions. We run Concorde and LKH in parallel for 32 instances on a 32 virtual CPU system (2Xeon E5-2620). Other results come from the original papers. The runtimes are reported for solving 10,000 test instances following \cite{kool2018attention}. \textbf{Bold} is the best among learning based methods, while `*' is the best in all methods. \par}
\caption{Multi-Decoder Attention Model (MDAM) vs Baselines}
\label{baseline}
\end{table*}

\section{Experiments}
In this section, we conduct experiments on six routing problems to verify the effectiveness of our method. Among them, TSP and CVRP are the most widely studied ones. TSP is defined as finding the shortest tour which visits each of the cities once and returns to the starting city, given the distances between each pair of the cities.
CVRP generalizes TSP, where the starting city must be a depot and every other city has a demand to be served by the vehicles. Multiple routes could be planned in CVRP, each for a vehicle, which visits a subset of cities with total demands not exceeding the capacity of the vehicle.
All the cities need to be covered by the routes. We follow existing works \cite{kool2018attention,nazari2018reinforcement} to generate instances with 20, 50 and 100 nodes (cities), which use two-dimensional Euclidean distance to calculate the distance between two cities, and the objective is to minimize the total travel distance. 
The coordinates of the city locations are sampled from the uniform distribution ranging from 0 to 1 for both dimensions independently. For CVRP, the vehicle capacities are fixed as 30, 40, 50 for problems with 20, 50, 100 nodes (cities), respectively. And the demands of each non-depot city are sampled from integers \{1...9\}.
Regarding the remaining four routing problems, i.e., Split Delivery Routing Problems (SDVRP), Orienteering Problem (OP) \cite{golden1987orienteering}, Prize Collecting TSP (PCTSP) \cite{balas1989prize} and Stochastic PCTSP (SPCTSP), the settings follow the existing work \cite{kool2018attention} and are introduced in the Supplementary Material. Note that these problems have their specific constraints and even random elements (SPCTSP). Nevertheless, MDAM is flexible enough to handle these properties by masking out the invalid nodes to visit at each step.

\noindent\textbf{Hyperparameters}. We embed the nodes with element-wise projection to 128-dimensional vectors. The Transformer encoder has 3 layers with 128 dimension features and 8 attention heads where the top one serves as the EG layer, and the hidden dimension of the fully connected layer (FF in Eq.~(\ref{eq102})) is 512. We choose the number of decoders in MDAM to be 5, and each of the 5 decoders takes 128 dimension vectors and 8 heads attention. For EG layer, we set the number of steps between re-embeddings to be 2, 4, 8 for TSP20, 50, 100 and 2, 6, 8 for CVRP20, 50, 100 for faster evaluation. Following \cite{kool2018attention}, we train the model with 2,500 iterations per epoch and batch size 512 (except 256 for CVRP100 to fit GPU memory constraint) for 100 epochs. We use Adam Optimizer \cite{kingma2014adam} with learning rate $10^{-4}$ for optimization. The coefficient of KL loss $k_{KL}$ needs to be large enough to keep the diversity between different decoders but not too large to deteriorate the performance of each decoder. We set $k_{KL}$ to 0.01 based on experiments on TSP20. Our code will be released soon \footnote{https://github.com/liangxinedu/MDAM}.

\begin{table*}[t] \small
\centering
\begin{tabular}{ll|rrr|rrr|rrr}
\toprule
\multicolumn{2}{c|}{\multirow{2}{*}{Method}} & \multicolumn{3}{c|}{n=20} & \multicolumn{3}{c|}{n=50} & \multicolumn{3}{c}{n=100} \\
\multicolumn{2}{c|}{}  & Obj & Gap & Time & Obj & Gap & Time & Obj & Gap & Time  \\
\midrule
\multirow{3}{*}{\rotatebox{90}{TSP}} & MDAM (no MD) greedy      & 3.85  & 0.27\% & 1s & 5.78 & 1.52\% & 3s & 8.06 & 3.76\% & 7s      \\
& MDAM (no EG) greedy & 3.84  & 0.06\% & 3s & 5.74 & 0.71\% & 8s & 7.98 & 2.78\% & 23s      \\
& MDAM greedy      & 3.84  & 0.05\% & 5s & 5.73 & 0.62\% & 15s & 7.93 & 2.19\% & 36s      \\
\midrule
\multirow{3}{*}{\rotatebox{90}{CVRP}} & MDAM (no MD) greedy      & 6.39  & 4.27\% & 1s & 10.93 & 5.32\% & 3s & 16.57 & 5.96\% & 9s      \\
& MDAM (no EG) greedy & 6.25  & 1.95\% & 4s & 10.79 & 3.95\% & 11s & 16.46 & 5.26\% & 26s      \\
& MDAM greedy      & 6.24  & 1.79\% & 7s & 10.74 & 3.47\% & 16s & 16.40 & 4.86\% & 45s      \\
\bottomrule
\end{tabular}
\caption{MDAM Structure Abaltion Results}
\label{MDAM_results}
\end{table*}

\subsection{Comparative Study}
Here we compare our method (MDAM) with existing strong deep learning based models. For testing, we sample 10,000 instances from the same distributions used for training.
To compute the optimality gap, we use the exact solver Concorde \cite{concorde2006} to get the objective values of the optimal solutions for TSP.
And for CVRP which is much harder to be solved exactly, we use the state-of-the-art heuristic solver LKH3 \cite{helsgaun2017extension} to get the benchmark solutions by following \cite{kool2018attention}.
For the beam search version of MDAM, we use 5 decoders each with beam size $B$=30 and 50 (denoted as bs30 and bs50), i.e. the whole beam size $\mathcal{B}$=150, and 250 respectively. Note that some methods (e.g. \cite{vinyals2015pointer}, \cite{khalil2017learning},  \cite{tsiligirides1984heuristic}, and \cite{deudon2018learning}) do not serve as baselines due to the reported inferior performance in \cite{kool2018attention}. We do not compare with L2I in \cite{iclr2020} either, due to its prohibitively long computation time. 
Without any instance parallelization, the average inference time of our model with 50-width beam search on one CVRP100 instance is 6.7s, while L2I needs 24 minutes (Tesla T4). For the other four problems, we compare MDAM with AM and other strong baselines as in \cite{kool2018attention}, with details introduced in Supplementary Material.

The results are summarized in Table \ref{baseline}. Note that all problems aim to minimize the objectives except OP, which aims to maximize the prizes collected along the tour. We can observe that for all the six routing problems with 20, 50 and 100 nodes, MDAM consistently outperforms AM \cite{kool2018attention} significantly. To avoid prohibitively long computation time for large problems, exact solver \citeauthor{gurobi} \shortcite{gurobi} is used as heuristics with time limits.
On CVRP and SDVRP, our MDAM with greedy decoding strategy significantly outperforms existing greedy decoding models in \cite{nazari2018reinforcement} and \cite{kool2018attention}, and the (standard) beam search version of RL \cite{nazari2018reinforcement}.
With our beam search scheme, MDAM outperforms not only the sampling version of AM \cite{kool2018attention}, but also the improvement heuristic NeuRewritter \cite{chen2019learning} on CVRP. Also, increasing beam size effectively further improves the solution quality for all the six problems at the cost of more inference time. For the comparison with traditional non-learning based methods, it is worth noting that MDAM outperforms sophisticated general purpose solvers OR Tools and Gurobi (with time limits) on large instances (OP and PCTSP with 100 nodes), and shows relatively good scalability.
For some problems such as OP with 50 nodes, MDAM outperforms highly specialized heuristic solvers.

A desirable property of construction heuristics is that, they can naturally handle uncertainty since the solutions are built incrementally. In contrast, optimization based methods (e.g. Gurobi, OR Tools and improvement heuristics) require some forms of online re-optimization (REOPT) \cite{kool2018attention} to adapt to the dynamic changes (e.g. constraint violation). For SPCTSP where the actual prize of each node is only known after visiting, MDAM significantly outperforms the REOPT strategies (details are in the Supplementary Material), AM greedy and sampling.

In terms of efficiency, though the baselines were executed on different machines, the computation time of MDAM is well accepted compared with existing deep learning based methods, especially considering the significant performance boost. Though MDAM (and all deep models) could be slower than the highly specialized solver on some problems (e.g. Concorde on TSP), it is generally much faster than the traditional algorithms with comparable solution quality.

\subsection{Ablation Study}

We further evaluate the effectiveness of different components in our model taking TSP and CVRP as the testbed. We assess the contribution of the Multi-Decoder structure (MD) and Embedding Glimpse layer (EG) to the quality of greedy construction policy, based on ablation study. We omit beam search here since it is an independent technique applicable to a given greedy decoding policy. The results are summarized in Table \ref{MDAM_results}.
We can observe that both MD and EG consistently improve the quality of learned construction policy for all instance sets,
which well verifies the rationale of our design. While MD boosts the performance significantly with relatively longer inference time, improvement by EG is relatively small but with little additional computation overhead.

\subsection{Analysis of MDAM}

\begin{figure}[t!]
        \centering
        \includegraphics[width=0.4\textwidth]{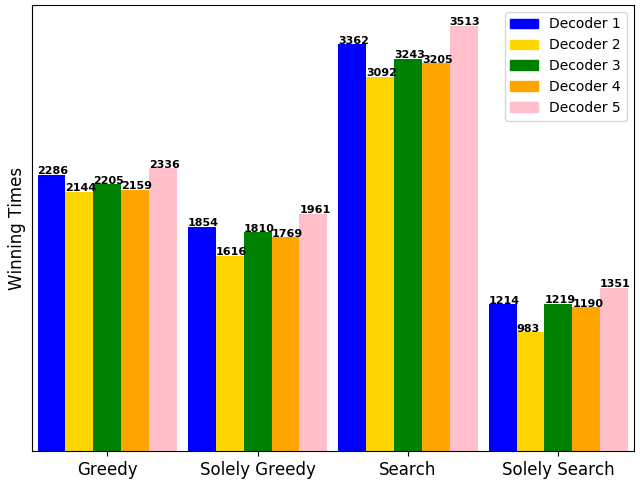}
        \caption{Greedy Decoding and Beam Search ($B$=50) Results of Each Decoder on CVRP20 }
        \label{f1}
    \end{figure}

To demonstrate the usefulness of the distinct construction patterns learned by each decoder, we evaluate the performance of each decoder on CVRP20 with greedy decoding and beam search ($B$=50). Figure \ref{f1} reports the number of times that each decoder finds the best (winning) and strictly best (solely winning) solution among all the 5 decoders in both the greedy decoding and search mode. We can see that it is common that more than one decoders find the best solution, since the winning times are higher than the solely winning times in both greedy decoding and beam search modes. More importantly, all decoders perform similarly since no one is dominant or not contributing at all, showing that each of them is indeed effective in the solving process.

In the Supplementary Material, we present more analysis of MDAM, including the effectiveness of the merging technique we designed for the customized beam search, and the impact of different number of decoders in the beam search mode.
We also compare with a stronger version of AM to further prove the effectiveness of MDAM. More specifically, we tune the temperature hyperparameter of the softmax output function of AM sampling to provide more diverse solutions. Finally, we show that comparing with AM, our MDAM has better generalization ability on larger problems such as CVRP with 150 and 200 nodes.

\section{Conclusions and Future Work}
In this paper, we propose a novel model to learn construction heuristics for routing problems, which is trained by reinforcement learning. It employs a multi-decoder structure to learn distinct construction patterns, which are further exploited by a customized beam search scheme. An Embedding Glimpse layer is incorporated in the model, which empowers the decoders with more informative embeddings. Our method outperforms state-of-the-art deep learning based methods on six routing problems, and engenders solutions close to traditional highly optimized solvers with reasonable time. In the future, we plan to improve our model by allowing flexibility in the number of decoders, and enabling collaborations among the decoders to make decisions jointly instead of individually.

\section{Acknowledgement}

The research was supported by the ST Engineering-NTU Corporate Lab through the NRF corporate lab@university scheme.
Part of this research was conducted at Singtel Cognitive and Artificial Intelligence Lab for Enterprises (SCALE@NTU), which is a collaboration between Singapore Telecommunications Limited (Singtel) and Nanyang Technological University (NTU) that is funded by the Singapore Government through the Industry Alignment Fund ‐ Industry Collaboration Projects Grant.
Wen Song was partially supported by the Young Scholar Future Plan of Shandong University (Grant No. 62420089964188). Zhiguang Cao was partially supported by the National Natural Science Foundation of China (61803104).

\bibliography{aaai}

\clearpage

\begin{strip}
\begin{center}
\textbf{\huge Supplementary Material}
\end{center}
\end{strip}

\renewcommand{\thesection}{}

\section{More details on the Comparative Study}

Below we introduce the four routing problems other than TSP and CVRP used in our experiments, and present the discussion on the corresponding experimental results as well. 

\paragraph{SDVRP.}
The Split Delivery Vehicle Routing Problem (SDVRP) is a variant of CVRP where the demand of a customer is allowed to be split and delivered over multiple routes. This problem is considered in \cite{kool2018attention} and \cite{nazari2018reinforcement}, hence we use these two methods as baselines. From the result in Table 1, we can see that when decoding greedily, MDAM significantly outperforms RL \cite{nazari2018reinforcement} and AM \cite{kool2018attention}. Moreover, MDAM with beam search significantly outperforms the beam search version of RL and sampling version of AM. These observations are consistent across different problem sizes.

\paragraph{OP.}
The Orienteering Problem (OP) \cite{golden1987orienteering} is to construct a tour starting and ending at the depot with the distance shorter than a threshold length. Every node is associated with a prize and the objective is to maximize the prize collected along the tour. Among the three variants used in \cite{kool2018attention}, we test on the hardest one where the prize of each node increases with the distance to the depot. Here, 
we compare with Gurobi \cite{gurobi} and a state-of-the-art Genetic Algorithm named Compass \cite{kobeaga2018efficient}. OR Tools \cite{google2016} and Tsili \cite{tsiligirides1984heuristic} are not presented here due to their relatively inferior performance as reported in \cite{kool2018attention}. 
Gurobi cannot find the optimal solutions for instances with 50 and 100 nodes due to the prohibitively long computation time, and can only give relatively poor solutions with the 30s time limit. Our MDAM gives the best solutions for n=50 while comes a bit short of Compass for n=100.

\paragraph{PCTSP.}
The Prize Collecting TSP (PCTSP) \cite{balas1989prize} is to construct a tour to collect prizes from nodes, while satisfying the constraint that the collected total prize should be equal to or higher than a threshold. The objective is to minimize the total tour length plus the sum of penalty associated with each unvisited nodes. For this problem, we compare our method with Gurobi (without and with 30s time limit), OR Tools (with 60s time limit), and a C++ version of Iterated Local Search (ILS) \cite{lourencco2003iterated} which is considered as the best open-source algorithm. These baseline solvers show good performances in the experiments, but take quite long computation time especially on large instances. In contrast, the learning based methods can give comparable solutions in much shorter time, and our MDAM outperforms AM on all instance sizes.

\paragraph{SPCTSP.}
The Stochastic Prize Collecting TSP (SPCTSP) is a variant of PCTSP with uncertainty, where the expectation of the prize associated with each node is known but the actual prize is only revealed upon visitation. Due to the uncertainty, a fixed tour planned before visitation could be invalid. A baseline algorithm named REOPT used in \cite{kool2018attention} plans a tour and executes part of it and then performs re-optimization using the C++ ILS algorithm. REOPT performs well when it either executes all node visits and plan additional node if the prize constraint is violated (REOPT all), or executes half of the nodes iteratively and then replans (REOPT half). From the results, we can see MDAM outperforms REOPT all and REOPT half by relatively large margins, which indicates that construction heuristics such as MDAM have their advantages in dealing with uncertainties.

\begin{table*}[!t] \small
\renewcommand\thetable{S.1} 
\centering
\begin{tabular}{ll|rrrrr|rr}
\toprule
 \multicolumn{2}{c|}{Method} & \multicolumn{5}{c|}{AM sampling (1,280 samples)} & MDAM (bs30) & MDAM (bs50) \\
 \midrule
\multirow{2}{*}{CVRP20} & Temp & 1 & 2 & 3 & 4 & 5 & - & - \\
& Obj & 6.25 & 6.19 & 6.18 & 6.19 & 6.26 & 6.14 & 6.14 \\
\midrule
\multirow{2}{*}{CVRP50} & Temp & 1 & 1.25 & 1.5 & 1.75 & 2 & - & - \\
& Obj & 10.62 & 10.61 & 10.60 & 10.59 & 10.60 & 10.50 & 10.48 \\
\midrule
\multirow{2}{*}{CVRP100} & Temp & 1 & 1.01 & 1.02 & 1.03 & 1.04 & - & - \\
& Obj & 16.23 & 16.1912 & 16.1909 & 16.1919 & 16.1925 & 16.03 & 15.99 \\
\bottomrule
\end{tabular}
\caption{MDAM Beam Searching with 30 and 50 solutions for each decoder vs AM Sampling with 1,280 solutions and various Softmax temperature}
\vspace{-5mm}
\label{temperature}
\end{table*}

\section{More Analysis of MDAM}
Here we provide more experiments and discussions from different angles to show the effectiveness of our approach.

\begin{figure}[t!]
    \renewcommand\thefigure{S.1} 
        \centering
        \includegraphics[width=0.32\textwidth]{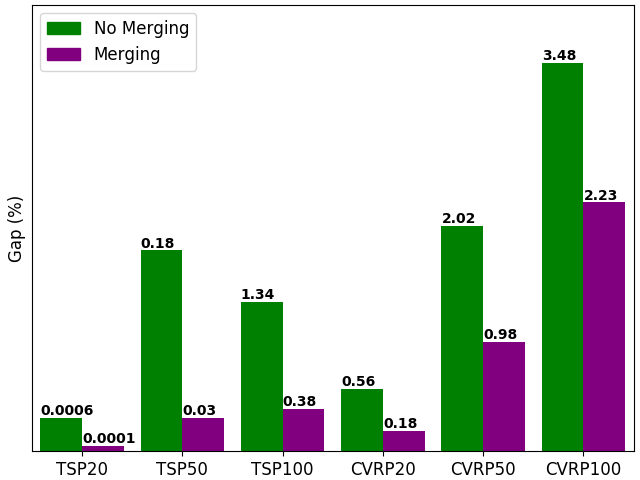}
        \caption{Results of 5 decoder beam search ($B$=50) with and without Merging}
        \label{f3}
        \vspace{-4mm}
    \end{figure}

\paragraph{Effectiveness of the merging technique.}
We compare the results of 5 decoder beam search ($B$=50) with and without merging, and plot the optimality gaps in Figure \ref{f3}. Clearly, merging significantly boosts the performance for all instance sets by eliminating inferior partial solutions. 
As stated before, ordinary beam searching could result in low variability and less diversity in solutions, which is likely why AM \cite{kool2018attention} uses sampling rather than beam searching. To verify this, we use CVRP100 as testbed.
The resulting objective value of AM \cite{kool2018attention} with beam search for 1,280 solutions on CVRP100 is 16.28, which is worse than AM sampling (1,280 solutions, obj=16.23) and our MDAM bs50 (250 solutions, obj=15.99).
To further demonstrate the effectiveness of our proposed model, we combine AM \cite{kool2018attention} beam searching and state merging as a baseline. AM with beam searching and state merging takes 145 and 227 minutes to achieve the objective values of 16.05 and 16.00 for width 150 and 250, respectively. In contrast, MDAM consumes much less time of 31 and 53 minutes to get better objective values of 16.03 and 15.99 for bs30 and bs50 (150 and 250 solutions in total). As we stated, merging state is computationally intensive for large beam due to the pairwise comparison. Hence we specifically design the customized beam search method for MDAM where the separate beam will not merge across different decoders.

\begin{figure}[t!]
    \renewcommand\thefigure{S.2} 
        \centering
        \includegraphics[width=0.32\textwidth]{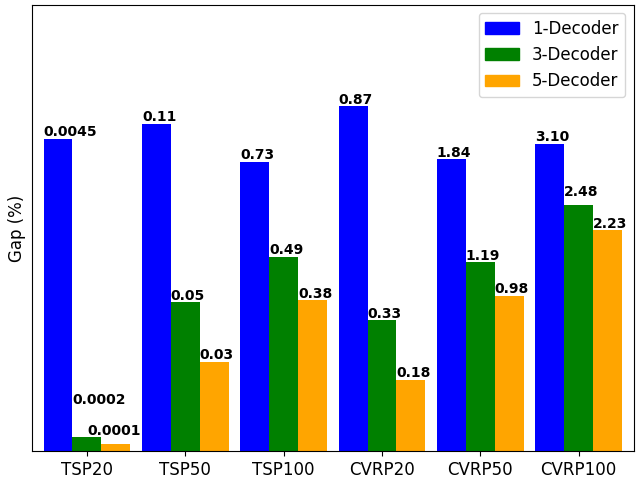}
        \caption{Results of beam search ($B$=50) with different number of decoders}
        \label{f2}
    \end{figure}

\paragraph{Impact of number of decoders in beam search.}
With the same beam size $B$=50, we compare the results of using 1, 3, and 5 decoders and plot the optimality gaps in Figure \ref{f2}. We can observe that the results are consistently better with more decoders, showing that more diverse construction patterns can indeed help to improve the performance of beam search. It also implies that the results could still be improved potentially with more than 5 decoders. Currently the number of decoders needs to be decided before training, which leaves an interesting future research direction to find more flexible ways to build variable number of decoders.

\paragraph{Comparing to AM with different temperature.}
The temperature hyperparameter in the softmax output function can affect the entropy of output probability for selecting nodes as well as the output solution diversity. Therefore we tune the temperature in the softmax output function of AM to provide stronger baselines. As shown in Table \ref{temperature}, tuning the temperature can provide better performance for AM sampling with 1,280 solutions. However, MDAM beam searching with beam size $B$=30 and 50 (denoted as bs30 and bs50) still outperforms AM sampling consistently by large margins. 
It is worth noting that CVRPs with more nodes need smaller temperature (closer to 1) to provide good performance.
The performance of AM sampling for large CVRP deteriorates fast with large temperature such as that AM sampling with temperature 2 has the objective value of 16.40 on CVRP100.

\begin{table}[!t] \small
\renewcommand\thetable{S.2} 
\centering
\begin{tabular}{l|rr|rr}
\toprule
 & \multicolumn{2}{c|}{n=150} & \multicolumn{2}{c}{n=200} \\
Method & Obj & Time & Obj & Time \\
\midrule
AM greedy & 23.52 & 1s & 30.23 & 2s \\
AM sampling (5) & 23.39 & 5s & 30.17 & 9s \\
AM sampling (1,280) & 22.87 & 30m & 29.54 & 49m \\
\midrule
MDAM greedy & 23.18 & 24s & 29.98 & 52s \\
MDAM bs30 & 22.66 & 13m & 29.37 & 30m \\
MDAM bs50 & 22.61 & 32m & 29.28 & 74m \\
\bottomrule
\end{tabular}
\caption{Generalization to larger problems}
\label{scale}
\vspace{-5mm}
\end{table}

\paragraph{Generalization Ability.}
In Table. \ref{scale}, we show the generalization ability of MDAM to larger problems taking CVRP as a demonstration.
Both MDAM and the baseline model AM are trained using CVRPs with 100 nodes and tested directly on CVRP150 and CVRP200.
On both problems, MDAM with greedy decoding strategy achieves better objective values than AM with greedy decoding and sampling with 5 solutions. At the same time, MDAM beam searching with beam size $B$=30 and 50 outperforms AM sampling with 1,280 solutions by large margins in comparable inferring time.

\end{document}